\documentclass[runningheads]{llncs}
\usepackage[T1]{fontenc}
\usepackage{graphicx}
\usepackage{subcaption}
\usepackage{booktabs}
\usepackage[misc]{ifsym}
\newcommand{\corr}{(\Letter)}
\usepackage[table]{xcolor}
\usepackage{mwe}

\begin{document}

\title{An Evaluation of Continual Learning for Advanced Node Semiconductor Defect Inspection}

\titlerunning{An Evaluation of Continual Learning for Advanced Node Semiconductor Defect Inspection}

\author{Amit Prasad\inst{1,*} \
Bappaditya Dey\inst{1,*} \corr \
Victor Blanco\inst{1} \and
Sandip Halder\inst{2,'}}

\authorrunning{Amit Prasad and Bappaditya Dey}

\institute{Interuniversity Microelectronics Centre, Kapeldreef 75, 3001, Belgium\inst{1} \\
SCREEN SPE Germany GmbH, Germany\inst{2} \\
Equal Contribution\inst{*} \\
This research was conducted during Sandip Halder’s tenure at imec\inst{'} \\
\email{amit.prasad.ext@imec.be, Bappaditya.Dey@imec.be}}

\maketitle              
\begin{abstract}
Deep learning-based semiconductor defect inspection has gained traction in recent years, offering a powerful and versatile approach that provides high accuracy, adaptability, and efficiency in detecting and classifying nano-scale defects. However, semiconductor manufacturing processes are continually evolving, leading to the emergence of new types of defects over time. This presents a significant challenge for conventional supervised defect detectors, as they may suffer from catastrophic forgetting when trained on new defect datasets, potentially compromising performance on previously learned tasks. An alternative approach involves the constant storage of previously trained datasets alongside pre-trained model versions, which can be utilized for (re-)training from scratch or fine-tuning whenever encountering a new defect dataset. However, adhering to such a storage template is impractical in terms of size, particularly when considering High-Volume Manufacturing (HVM). Additionally, semiconductor defect datasets, especially those encompassing stochastic defects, are often limited and expensive to obtain, thus lacking sufficient representation of the entire universal set of defectivity. This work introduces a task-agnostic, meta-learning approach aimed at addressing this challenge, which enables the incremental addition of new defect classes and scales to create a more robust and generalized model for semiconductor defect inspection. We have benchmarked our approach using real resist-wafer SEM (Scanning Electron Microscopy) datasets for two process steps, ADI and AEI, demonstrating its superior performance compared to conventional supervised training methods.
\end{abstract}

\section{Related Work}
In the semiconductor process (mainly, Litho-Etch) domain, numerous approaches have been suggested for defect classification and localisation \cite{9971022}, \cite{10.1117/12.2622550}, \cite{10.1117/12.2675573}. To the best of the authors' knowledge, the concept of incremental learning \cite{masana2022classincremental} for multi-class, multi-instance defect detection on SEM images has previously not been explored.

\section{Methodology}
\subsection{Dataset}
\label{sec:dataset}
\begin{figure}
     \centering
     \begin{subfigure}[b]{0.49\textwidth}
         \centering
         \includegraphics[width=\textwidth]{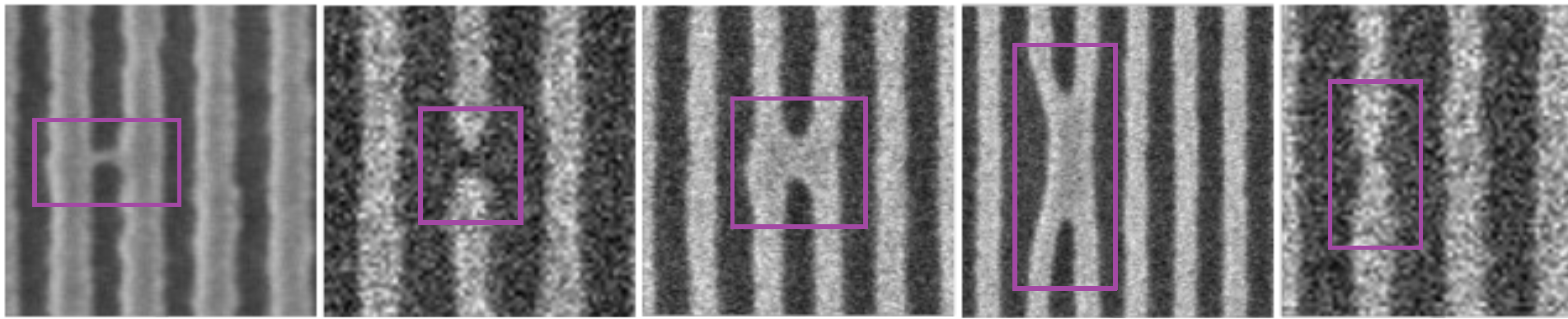}
         \caption{ADI defects. Left to right: Microbridge, Gap, Bridge, Line collapse, Probable-gap}
        \label{fig:adi_defects}
     \end{subfigure}
     \hfill
     \begin{subfigure}[b]{0.49\textwidth}
         \centering
         \includegraphics[width=\textwidth]{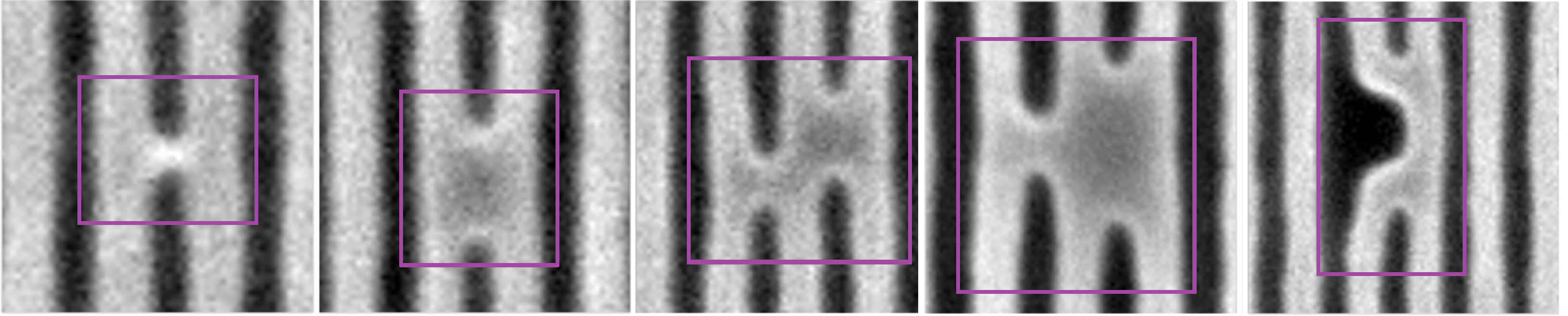}
         \caption{AEI defects. Left to right: Thin bridge, Single bridge, Multi bridge non-horizontal, Multi bridge horizontal, Line collapse}
         \label{fig:aei_defects}
     \end{subfigure}
     \hfill
     \setlength{\belowcaptionskip}{-15pt} 
        \caption{SEM images with a) ADI defects and b) AEI  defects}
        \label{fig:defect_images}
\end{figure}

Original (resist) wafer SEM (Scanning Electron Microscopy) images were obtained during ADI (After Development Inspection) and AEI (After Etch Inspection) stages. Figure \ref{fig:defect_images} illustrates exemplary defect types in both process steps. 
The instance distribution per defect class is captured in Table \ref{tab:dataset}. 
\vspace{-12pt}
\begin{table}[htbp]
  \centering
  \caption{Instance distribution per class}
  \resizebox{1\linewidth}{!}
  {
    \begin{tabular}{cccccr|ccccc}
          &       & \multicolumn{3}{c}{\textbf{ADI Instances}} &       &       &       & \multicolumn{3}{c}{\textbf{AEI Instances}} \\
    \textbf{Label} & \textbf{Defect type} & \textbf{Training} & \textbf{Validation} & \textbf{Test} &       & \textbf{Label} & \textbf{Defect type} & \textbf{Training} & \textbf{Validation} & \textbf{Test} \\
    0     & \textbf{Microbridge} & 380   & 47    & 78    &       & 5     & \textbf{Thin bridge} & 241   & 29    & 29 \\
    1     & \textbf{Gap} & 1046  & 156   & 174   &       & 6     & \textbf{Single bridge} & 240   & 29    & 31 \\
    2     & \textbf{Bridge} & 238   & 19    & 17    &       & 7     & \textbf{Multi bridge non-horizontal} & 160   & 21    & 19 \\
    3     & \textbf{Line collapse} & 550   & 66    & 76    &       & 8     & \textbf{Multi bridge Horizontal} & 80    & 10    & 10 \\
    4     & \textbf{Probable-gap} & 315   & 49    & 54    &       & 9     & \textbf{Line collapse} & 202   & 40    & 34 \\
\cmidrule{1-5}\cmidrule{7-11}          & Total & 2529  & 337   & 399   &       &       & Total & 923   & 129   & 123 \\
    \end{tabular}%
    }
  \label{tab:dataset}%
\end{table}%
\vspace{-15pt}
\subsection{Notations and Preliminaries}
The following notations have been used in this work.
\begin{definition}
\textbf{Task ($T_p$)}: This is defined as supervised training of a defect detection framework for $p$ classes (0 to $p-1$) in the dataset of the form $(x_i, y_i)_{i=1}^m$ (m instances with defect feature $x_i$and corresponding label $y_i$). This is denoted by $T_p$.
\end{definition}
\begin{definition}
\textbf{Finetuned task ($F_p^q$)}: This is defined as supervised training of a defect detection framework for next $q$ classes ($p$ to $q-1$) in the dataset of the form $(x_i, y_i)_{i=1}^m$, which has previously been trained on the initial $p$ classes (0 to $p-1$). However, it's important to note that identifying these initial $p$ classes is not guaranteed.This is denoted by $F_p^q$.
\end{definition}
\begin{definition}
\textbf{Incremental task ($\mathcal{T}_p^q$)}: This is defined as \textbf{incremental} supervised training of a defect detection framework for next $q$ classes ($p$ to $q-1$) in the dataset of the form $(x_i, y_i)_{i=1}^m$, which has previously been trained on the initial $p$ classes (0 to $p-1$), enabling it to identify all $(p+q)$ classes. This is denoted by $\mathcal{T}_p^q$.
\end{definition}
\subsection{Structure of study}
In this work we present the following case studies.
\begin{enumerate}
\item \textbf{Case study 1} (see Section \ref{sec:case_study_1}) examines effectiveness of the framework in \textbf{incrementally} learning new defect classes and minimizing \textbf{forgetting} of previously trained defect classes on the ADI dataset.
\item \textbf{Case study 2} (see Section \ref{sec:case_study_2}) 
assesses framework for \textbf{incrementally} learning new defect classes in AEI images and minimizing \textbf{forgetting} of previously trained defect classes across the entire ADI dataset.
\item \textbf{Case study 3} (see Section \ref{sec:case_study_3}) compares three training strategies: (i) conventional supervised training strategy with all defect classes at once,
(ii) conventional supervised training with first $p$ defect classes and then fine-tune on new $q$ defect classes,
(iii) proposed \textbf{incremental} supervised training strategy with first $p$ defect classes and then fine-tune on new $q$ defect classes.
\end{enumerate}
We use the Faster-RCNN \cite{10.5555/2969239.2969250} model for all studies. Moreover, for incremental tasks, the approach utilized is presented in \cite{9599446} which also uses FRCNN.
\section{Case study 1}
\label{sec:case_study_1}
\begin{figure}
     \centering
     \begin{subfigure}[b]{0.49\textwidth}
         \centering
         \includegraphics[width=\textwidth]{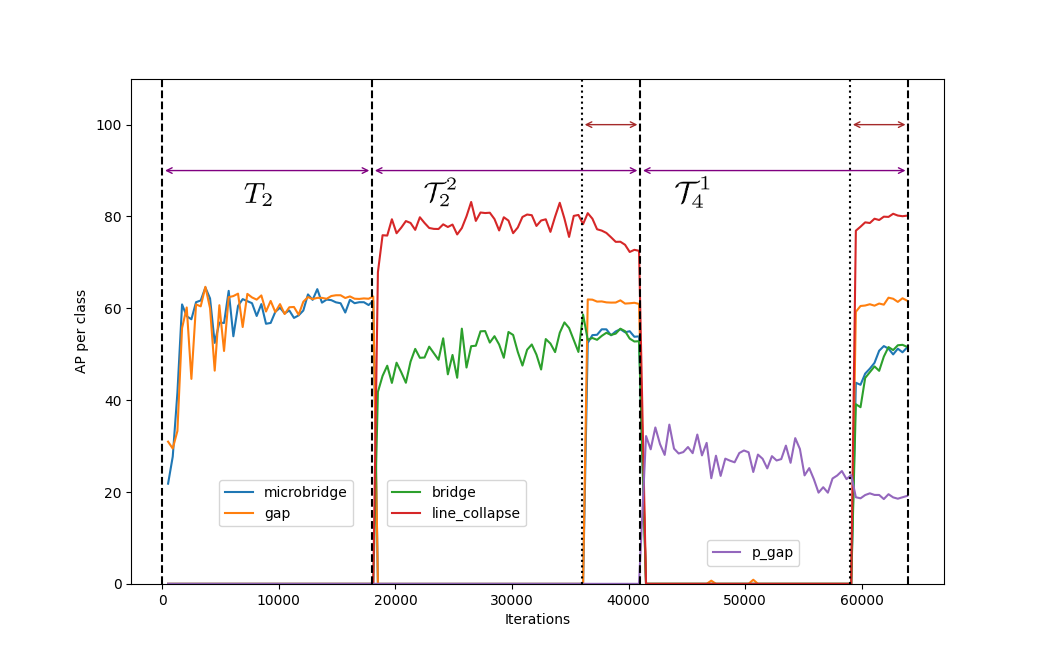}
         \caption{Model trained incrementally for tasks $\mathcal{T}_2^2$ and $\mathcal{T}_4^1$ after training on task $T_2$}
         \label{fig:cl_2_p_2_p_1}
     \end{subfigure}
     \hfill
     \begin{subfigure}[b]{0.49\textwidth}
         \centering
         \includegraphics[width=\textwidth]{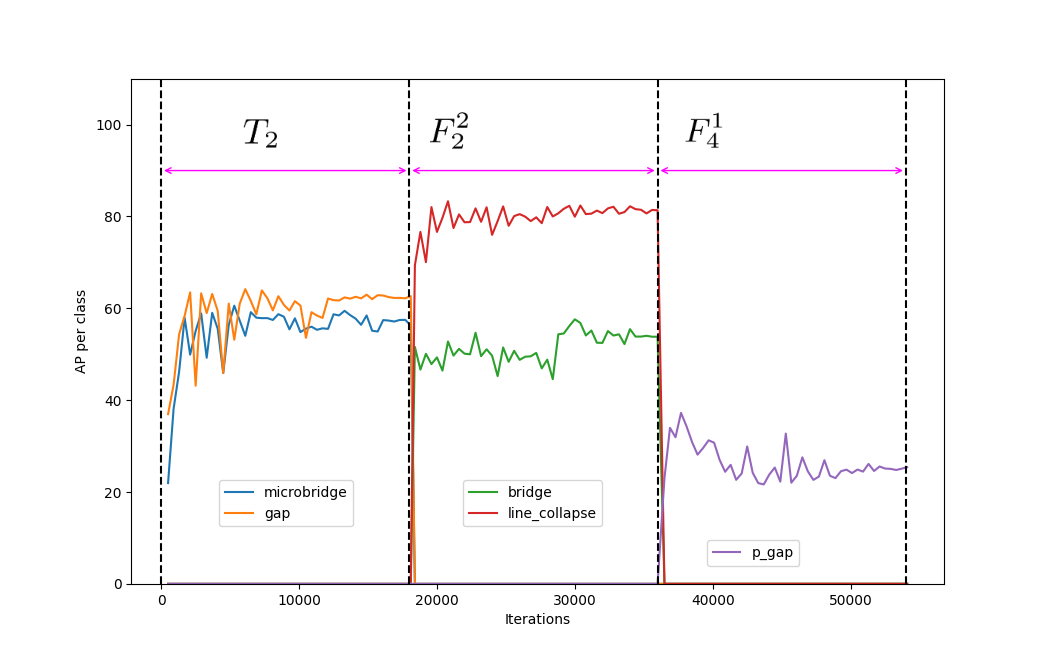}
         \caption{Model finetuned on tasks $F_2^2$ and $F_4^1$ after task $T_2$}
         \label{fig:ncl_2_p_2_p_1}
     \end{subfigure}
     \hfill
     \setlength{\belowcaptionskip}{-15pt} 
        \caption{Comparison between (a) proposed incremental learning and (b) conventional fine-tuning method.}
        \label{fig:2_p_2_p_1}
\end{figure}

The model starts training with the task $T_2$ (initially trained for 2 defect classes, microbridge and gap), followed by two consecutive incremental training tasks: $\mathcal{T}_2^2$ (adding 2 more defect classes, bridge and line-collapse), and finally $\mathcal{T}_4^1$ (adding the last defect class as probable gap), using the ADI dataset. For an evaluation of performance, average precision (AP) per defect class vs iterations is plotted, marking checkpoints where new defect classes were introduced and where continual learning takes place. The results are compared to the conventional fine-tuning approach, where the model has been trained on tasks $F_2^2$ and $F_4^1$, while keeping all experimental conditions constant. In Figure \ref{fig:2_p_2_p_1} a), it is evident how effective incremental learning is for progressively learning defect classes and minimizing catastrophic forgetting. Conversely, in Figure \ref{fig:2_p_2_p_1} b), it is apparent how swiftly catastrophic forgetting occurs in the case of fine-tuning.
\section{Case study 2}
\label{sec:case_study_2}
\begin{figure}
     \centering
     \begin{subfigure}[b]{0.49\textwidth}
         \centering
         \includegraphics[width=\textwidth]{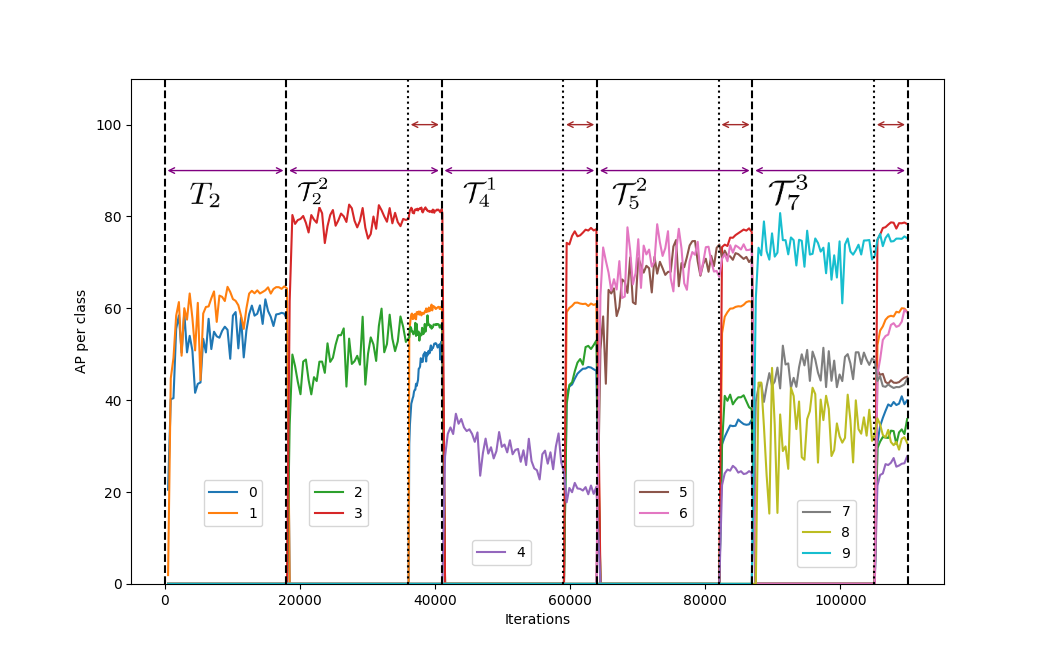}
         \caption{Model trained incrementally for tasks $\mathcal{T}_2^2$, $\mathcal{T}_4^1$, $\mathcal{T}_5^2$, $\mathcal{T}_7^3$ after training on task $T_2$}
         \label{fig:cl_adi_aei}
     \end{subfigure}
     \hfill
     \begin{subfigure}[b]{0.49\textwidth}
         \centering
         \includegraphics[width=\textwidth]{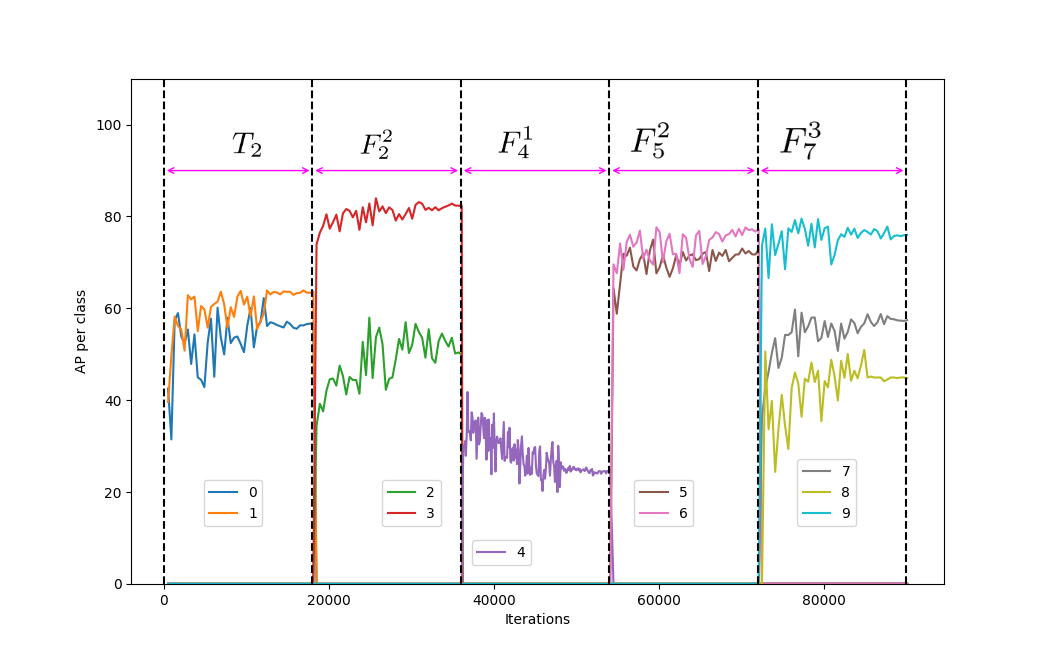}
         \caption{Model finetuned on tasks $F_2^2$ and $F_4^1$, $F_5^2$ and $F_7^3$ after task $T_2$}
         \label{fig:nc_adi_aei}
     \end{subfigure}
     \hfill
        \setlength{\belowcaptionskip}{-30pt} \caption{(a) Proposed incremental learning vs (b) conventional fine-tuning method for incremental learning of AEI defects, after training across the ADI dataset.}
        \label{fig:adi_aei}
        
\end{figure}

\vspace{-2pt}
Defect classes from the AEI dataset are incrementally added following training on the ADI dataset. The model, following task $\mathcal{T}_4^1$, undergoes training on tasks $\mathcal{T}_5^2$ and $\mathcal{T}_7^3$. Similarly, following task $F_4^1$, the model undergoes fine-tuning for tasks $F_5^2$ and $F_7^3$. The Figure \ref{fig:adi_aei} illustrates the comparison between proposed incremental learning and conventional fine-tuning (using AP vs iteration plot).
\vspace{-8pt}
\section{Case study 3}
\label{sec:case_study_3}
\begin{figure}[!htb]
  \centering
  \includegraphics[scale=0.45]{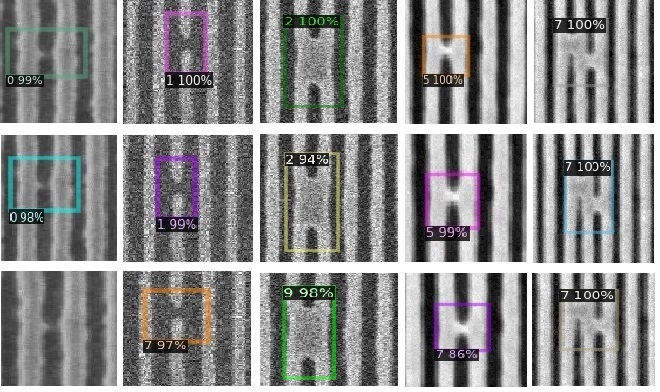}
  \caption{Upper row: Model trained for defect detection on all classes at once.\\Middle row: Model obtained after incremental training on task $T_7^3$. \\Lower row: Model obtained after training on task $F_7^3$\\Defect types (ground truth), left to right: Microbridge, Gap, Bridge, Thin bridge, Multi bridge non-horizontal.}
  \label{fig: cl_vs_ncl_pic}
\end{figure}

\vspace{-6pt}
Inference results are shown in Figure \ref{fig: cl_vs_ncl_pic} (with corresponding labels, bounding boxs and confidence scores) are from 3 training strategies, first is the model trained on task $T_{10}$ (incorporating all defect classes simultaneously) while the other models are derived from tasks $\mathcal{T}_7^3$ and $F_7^3$. The labels are referenced from Table \ref{tab:dataset}. Notably, it is observed that the model after task $T_7^3$ performs comparably to the model trained on task $T_{10}$. However, the model obtained after task $F_7^3$ demonstrates forgetfulness or mislabeling of defects it encountered earlier, as it has only recently been exposed to labels 7, 8, and 9.
\vspace{-15pt}
\section{Conclusion}
In this study, we demonstrated the effectiveness of a continual learning strategy in progressively learning the classification and localization of semiconductor defect classes in aggressive pitches, while mitigating catastrophic forgetting.
%
%
%
\bibliography{sn-bibliography}
\end{document}